\documentclass[sigconf, screen]{acmart}

\renewcommand\footnotetextcopyrightpermission[1]{}
\usepackage{makecell}
\usepackage{cleveref}
\usepackage{diagbox}
\usepackage{colortbl}

\AtBeginDocument{
  }

\begin{document}

\fancyhf{}

\title{Towards Generalized and Training-Free Text-Guided Semantic Manipulation}

\newcommand{\methodname}{GTF}


\author{Yu Hong}
\affiliation{
  \institution{UESTC$ ^1 $}
  \thanks{1 University of Electronic Science and Technology of China}
  \country{}}
\email{ayanamiyuk@gmail.com}

\author{Xiao Cai}
\affiliation{
  \institution{UESTC$ ^1 $} \country{}}
\email{xiaocai628@gmail.com}

\author{Pengpeng Zeng}
\affiliation{
  \institution{Tongji University} \country{}}
\email{is.pengpengzeng@gmail.com}

\author{Shuai Zhang}
\affiliation{
  \institution{UESTC$ ^1 $} \country{}}
\email{}

\author{Jingkuan Song}
\affiliation{
  \institution{Tongji University} \country{}}
\email{}

\author{Lianli Gao}
\affiliation{
  \institution{UESTC$ ^1 $} \country{}}
\email{}

\author{Heng Tao Shen}
\affiliation{
  \institution{Tongji University} \country{}}
\email{}

\renewcommand{\shortauthors}{Hong et al.}

\begin{abstract}

Text-guided semantic manipulation refers to semantically editing an image generated from a source prompt to match a target prompt, enabling the desired semantic changes (e.g., addition, removal, and style transfer) while preserving irrelevant contents. With the powerful generative capabilities of the diffusion model, the task has shown the potential to generate high-fidelity visual content. Nevertheless, existing methods either typically require time-consuming fine-tuning (inefficient), fail to accomplish multiple semantic manipulations (poorly extensible), and/or lack support for different modality tasks (limited generalizability). Upon further investigation, we find that the geometric properties of noises in the diffusion model are strongly correlated with the semantic changes. Motivated by this, we propose a novel \textbf{\emph{\methodname}} for text-guided semantic manipulation, which has the following attractive capabilities: 1) \textbf{Generalized}: our \emph{\methodname} supports multiple semantic manipulations (e.g., addition, removal, and style transfer) and can be seamlessly integrated into all diffusion-based methods (i.e., Plug-and-play) across different modalities (i.e., modality-agnostic); and 2) \textbf{Training-free}: \emph{\methodname} produces high-fidelity results via simply controlling the geometric relationship between noises without tuning or optimization. Our extensive experiments demonstrate the efficacy of our approach, highlighting its potential to advance the state-of-the-art in semantics manipulation. Our project page is at \href{https://ayanami-yu.github.io/GTF-Project-Page/}{https://ayanami-yu.github.io/GTF-Project-Page/}.

\end{abstract}  


\begin{teaserfigure}
  \includegraphics[width=\textwidth]{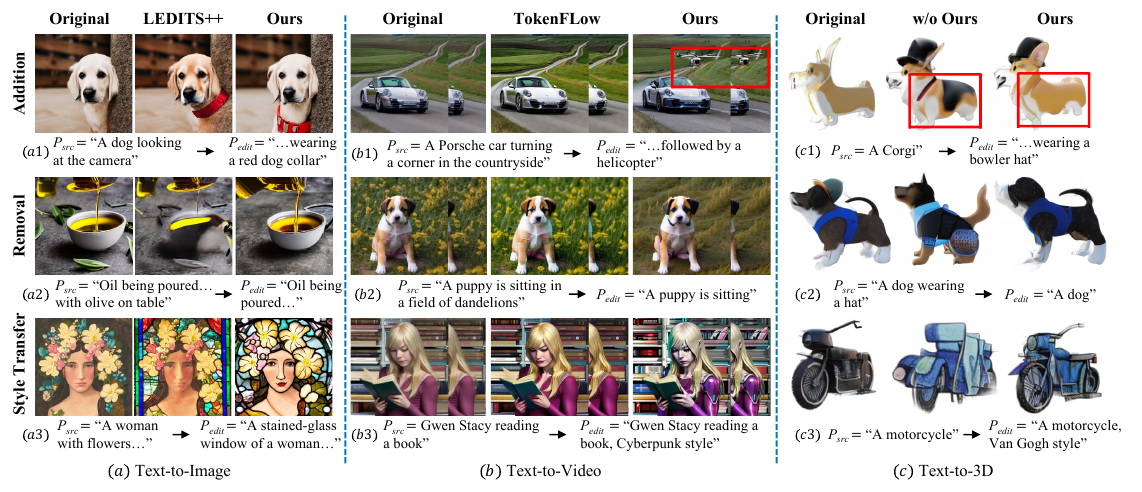}
  \vspace{-24pt}
  \caption{Result Comparison between the state-of-the-art method and our proposed method (\textbf{\emph{\methodname}} ) for text-guided semantic manipulation across different modalities, including image (a), video (b), and 3D (c). From the figure, we can observe that \textbf{\emph{\methodname}} supports multiple semantics manipulations and can be seamlessly integrated into all diffusion-based methods, i.e., Plug-and-play, producing high-fidelity visual results. \( P_{\text{src}} \) and \( P_{\text{edit}} \) denote the source prompt and the edited prompt, respectively.
  }
  \Description{}
  \label{fig:teaser}
\end{teaserfigure}

\maketitle

\section{Introduction}

Text-guided semantic manipulation~\cite{kawar2023imagictextbasedrealimage, meng2022sdeditguidedimagesynthesis, kim2022diffusioncliptextguideddiffusionmodels} modifies a user-generated image based on a source prompt to match a target prompt that involves the user's desired semantic changes (e.g., addition, removal, and style transfer) while maintaining minimal visual changes. This approach can be applied to multiple domains, such as photography, advertising, and social media. As shown in \cref{fig:teaser} (a1),  the source image \emph{I} is generated with the source prompt ``a dog looking at the camera'' and then \emph{I} is edited with the target prompt ``A dog looking at the camera wearing a red dog collar''. In addition to involving the image modality, the task has been spawned into other modalities, including video and 3D, as illustrated in \cref{fig:teaser} (b) and (c), respectively. Till now, it is still challenging to change the local semantic information without affecting the original visual information, no matter what modality. 

For the text-guided semantic manipulation task, it requires a model capable of first generating high-quality visual content based on a source prompt, and secondly, accomplishing semantic editing of the final generated target content based on a target prompt. With the availability of a massive amount of text-visual
paired data and large-scale vision-language models, diffusion models have demonstrated outstanding generation capabilities in this area~\cite{Rombach_2022_CVPR, esser2024scalingrectifiedflowtransformers}.
Despite sharing a common generative backbone, existing methods for text-guided semantic manipulation are predominantly modality-specific.
For images~\cite{hertz2022prompt, cao_2023_masactrl, brack2024leditslimitlessimageediting}, approaches often rely on cross-attention modulation, noise inversion, or prompt interpolation to steer generation while maintaining structural fidelity.
For videos~\cite{qi2023fatezero, yang2023rerendervideozeroshottextguided, geyer2023tokenflowconsistentdiffusionfeatures}, semantic editing is typically conducted either by applying frame-wise image editing followed by temporal consistency regularization, or by integrating spatiotemporal attention to enforce coherence across frames.
For 3D content~\cite{li2024instructpix2nerfinstructed3dportrait, haque2023instructnerf2nerfediting3dscenes, palandra2024gseditefficienttextguidedediting}, manipulation is commonly realized by embedding textual semantics into the optimization of neural 3D representations (e.g., 3DGS~\cite{kerbl20233dgaussiansplattingrealtime}, NeRF~\cite{mildenhall2020nerfrepresentingscenesneural}). While effective within individual domains, these strategies lack generality and are not readily transferable across modalities.
Thus, this raises an intuitive question: \emph{since existing mainstream approaches are based on diffusion models, can a generalized method be devised to achieve high-fidelity visual content across different modalities?}

With this question in mind, we revisit the generative process of diffusion models under different prompts. Specifically, when conditioned on a source prompt \( P_{\text{src}} \) and a target prompt \( P_{\text{tgt}} \), the diffusion model yields two distinct noise trajectories that reflect their respective semantic intents. Notably, such prompts often share partial semantics (e.g., “a cat” vs. “a cat with glasses”), suggesting a potential corresponding overlap in their noise distributions.
If such a semantic correspondence indeed manifests in the noise space, then directly manipulating the predicted noise vectors offers a principled pathway for controllable generation. Supporting this intuition, prior work~\cite{song2020generativemodelingestimatinggradients} reveals that the noise predicted by diffusion models approximates the gradient of the log-likelihood, effectively encoding a direction in semantic space. This observation provides a theoretical foundation for treating noise-level operations as a viable mechanism for semantic control.

Inspired by the insights presented above, we further theoretically investigate the feasibility of semantic manipulation through noise control in diffusion models. Building upon this perspective, we categorize semantic editing into two fundamental operations — semantic addition and semantic removal — and design dedicated noise composition strategies tailored to each.  
These strategies are integrated into \textbf{GTF}, which achieves controllable editing by explicitly combining the noise vectors predicted under the source and target prompts. Specifically, at each denoising step, GTF computes a weighted combination of the source and target conditional noise to steer the diffusion trajectory toward the desired semantic direction. This flexible design enables GTF to plug into existing diffusion backbones without retraining or architecture modification.

To summarize, our main contributions are threefold:
\begin{itemize}
    \item  We reveal that the geometric structure of noise in diffusion models inherently encodes semantic directions, providing a foundation for controllable semantic manipulation.
    \item We propose \textbf{GTF}, a \textbf{G}eneralized and \textbf{T}raining-\textbf{F}ree pipeline for accurate and efficient semantic manipulation across multiple modalities and tasks.
    \item Extensive experiments demonstrate the strong controllability and generalizability of our approach, highlighting its effectiveness and potential to advance the field of diffusion-based semantic manipulation.
\end{itemize}

\section{Related Works}

\noindent\textbf{Text-to-Image Manipulation.}
Text-to-image editing has evolved from GAN-based methods \cite{dhamo2020semanticimagemanipulationusing, Frolov_2021, xia2022ganinversionsurvey} to diffusion-based approaches \cite{ho2020denoisingdiffusionprobabilisticmodels, song2022denoisingdiffusionimplicitmodels}, which offer improved cross-modal consistency and generation quality. Techniques such as SDEdit \cite{meng2022sdeditguidedimagesynthesis} and DiffusionCLIP \cite{kim2022diffusioncliptextguideddiffusionmodels} enable guided editing through noise manipulation or CLIP-based supervision. Methods like DreamBooth \cite{ruiz2023dreamboothfinetuningtexttoimage}, Textual Inversion \cite{gal2022imageworthwordpersonalizing}, and Imagic \cite{kawar2023imagictextbasedrealimage} further enhance control via prompt tuning or model fine-tuning.

\noindent\textbf{Text-to-Video Manipulation.}
Text-driven video editing poses challenges in maintaining temporal consistency. Early models \cite{molad2023dreamixvideodiffusionmodels, esser2023structurecontentguidedvideosynthesis} relied on full-model training, while recent works adopt pre-trained T2V models for efficiency \cite{chen2024controlavideocontrollabletexttovideodiffusion, yan2023magicpropdiffusionbasedvideoediting}. Methods like Pix2Video \cite{ceylan2023pix2videovideoeditingusing} and Vid2vid-zero \cite{wang2024zeroshotvideoeditingusing} improve edit propagation and coherence. Others, such as VidEdit \cite{couairon2024videditzeroshotspatiallyaware} and Edit-A-Video \cite{shin2023editavideosinglevideoediting}, achieve object-level precision through spatial awareness and attention mechanisms. Optical flow-based techniques \cite{chu2023medmmediatingimagediffusion, cong2024flattenopticalflowguidedattention} further enhance temporal stability.

\noindent\textbf{Text-to-3D Manipulation.}
Text-to-3D editing has progressed from image-based reconstruction \cite{Oppenlaender_2022, haque2023instructnerf2nerfediting3dscenes} to multi-view consistent editing \cite{dong2024vicanerfviewconsistencyaware3dediting, fang2023editing3dscenestext} and general-purpose frameworks \cite{khalid2024latenteditortextdrivenlocal}. Diffusion models have been adapted for 3D generation via pre-trained 2D models \cite{poole2022dreamfusiontextto3dusing2d, chen2023shapeditorinstructionguidedlatent3d}, with improvements in geometric consistency and text-vision alignment \cite{ma2023xmeshfastaccuratetextdriven, hyung2023local3dediting3d}. Recent advances in SDS enable fine-grained 3D editing for complex objects and scenes \cite{cheng2024progressive3dprogressivelylocalediting, mikaeili2023skedsketchguidedtextbased3d}.

\section{Methodology}
In this section, we introduce \textbf{GTF}, a generalized and training-free framework for text-guided semantic manipulation. GTF enables controllable addition and removal of semantics in diffusion-based multimodal generative models. We begin by revisiting essential preliminaries to provide the theoretical grounding for our approach.

\begin{figure*}
    \centering
    \includegraphics[width=\linewidth]{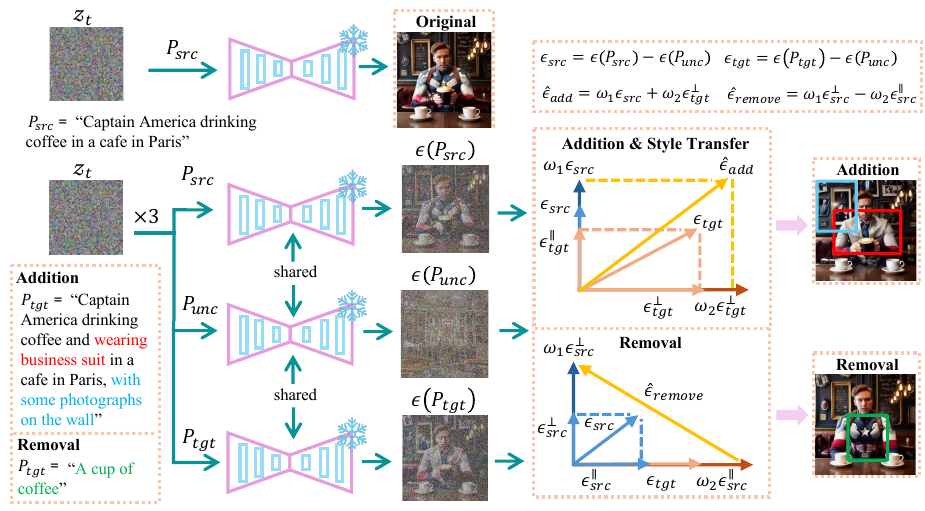}
    \vspace{-20pt}
    \caption{\textbf{Overview of the GTF pipeline.} Given a pair of source and target prompts, we aim to perform semantic addition or removal by combining their corresponding noise predictions. Specifically, at each diffusion step, we first predict the unconditional noise, followed by two conditional noise predictions based on the source and target prompts, respectively. Depending on the manipulation type (addition or removal), we apply our algorithm to combine these noises and compute the final guidance noise for the current step. This guidance is applied iteratively at every denoising step, progressively steering the generation toward the edited result.}
    \Description{}
    \label{fig:pipeline}
    \vspace{-6pt}
\end{figure*}

\subsection{Preliminaries: Latent Diffusion Models}

Latent Diffusion Models (LDMs)~\cite{Rombach_2022_CVPR} perform denoising in a compressed latent space encoded by a pretrained autoencoder, significantly improving computational efficiency while preserving visual quality. The generation process involves a forward noising phase and a reverse denoising phase, progressively transforming noise into semantically meaningful content.

In conditional settings, LDMs incorporate text embeddings to guide noise prediction at each step, enabling controllable generation. The model learns to reconstruct the original latent by minimizing the error between predicted and true noise, establishing a foundation for text-driven semantic manipulation.

\subsection{Exploring Semantic Manipulation in Noise Predictions}
\label{sec:explore}

Diffusion models have shown remarkable success in conditional generation tasks, yet their underlying mechanism can also be interpreted through the lens of score-based generative modeling. Let \( p_{\text{data}}(x) \) denote the distribution of real images, and let \( p_{\text{data}}(x \mid c) \) represent the conditional distribution given a text prompt \( c \). The goal of conditional generation is to sample from \( p_{\text{data}}(x \mid c) \), i.e., to generate outputs consistent with the conditioning semantics.

The score function, defined as the gradient of the log-probability \( \nabla_x \log p_{\text{data}}(x \mid c) \), points toward higher-density regions of the conditional distribution. As illustrated in SMLD~\cite{song2020generativemodelingestimatinggradients}, the noise prediction \( \epsilon_\theta(x_t, c) \) approximates the score function of the perturbed distribution:
\begin{equation}
    \epsilon_\theta(x_t, c) \approx -\sigma_t \cdot \nabla_x \log p_t(x_t \mid c),
    \label{eq:noise_pred}
\end{equation}
where \( \sigma_t \) is the standard deviation of noise at timestep \( t \), and \( p_t(x_t \mid c) \) denotes the noisy conditional data distribution.

This interpretation reveals that the predicted noise vector inherently encodes a semantic direction in the data space toward more probable samples under condition \( c \). Therefore, geometric operations on noise vectors—such as directional addition or subtraction—can be understood as explicit manipulations of the generative trajectory within the learned conditional score field.

\noindent\textbf{Semantic Addition.} 
Now consider a scenario where semantic addition is required, and let \( c_1 \) and \( c_2 \) denote prompt conditions (e.g., source and target semantics). Each prompt induces its own conditional distribution over \( x \). Assuming conditional independence given \( x \), the joint conditional distribution can be derived via Bayes' rule as:
\begin{equation}
\begin{split}
p(x \mid c_1, c_2) & = \frac{p(x, c_1, c_2)}{p(c_1, c_2)} 
= \frac{p(x) \cdot \prod_{i=1}^2 p(c_i \mid x)}{\prod_{i=1}^2 p(c_i)} \\
& = \frac{p(x) \cdot \prod_{i=1}^2 \frac{p(x \mid c_i) \cdot p(c_i)}{p(x)}}{\prod_{i=1}^2 p(c_i)}
\end{split}
\label{eq:combine_prob}
\end{equation}

Then we differentiate \cref{eq:combine_prob} with respect to $ x $ after taking the logarithm and compute as:
\begin{equation}
\begin{split}
\nabla_x \log p(x \mid c_1, c_2) 
&= \nabla_x \log p(x) 
\\&+ \sum_{i=1}^2[\nabla_x \log p(x \mid c_i) - \nabla_x \log p(x)] 
\end{split}
\label{eq:log_prob}
\end{equation}

Recall \cref{eq:noise_pred}, we can transform \cref{eq:log_prob} to derive the final combined guidance noise \( \epsilon_{\text{pred}} \) as:
\begin{equation}
    \epsilon_{\text{pred}} = \epsilon(\varnothing) + \sum_{i=1}^2[\epsilon(c_i)-\epsilon(\varnothing)],
    \label{eq:pred_add}
\end{equation}
where \( \epsilon(\varnothing) \) denotes the unconditional noise prediction. 
This formulation illustrates that the semantic addition task can be effectively realized via a combination of individual semantic noise components.

\noindent\textbf{Semantic Removal.}
Similarly, for semantic removal, we treat the combined prompt \( \{c_1, c_2\} \) as the source prompt and \( c_2 \) as the semantic element to be removed, i.e. target prompt. To isolate the remaining semantics in \( c_1 \), we derive the guidance noise as follows:
\begin{equation}
\begin{split}
p(x \mid c_1) & = \frac{p(x)\cdot p( c_1 \mid x)}{p(c_1)} 
= \frac{p(x) }{p(c_1)} \cdot \frac{p(c_1,c_2 \mid x) }{ p(c_2 \mid x)}\\
& 
= \frac{p(x) }{p(c_1)} \cdot \frac{p(c_1,c_2) \cdot \frac{p(x\mid c_1, c_2)}{p(x)} }{ p(c_2) \cdot \frac{p(x \mid c_2)}{p(x)}}\\
& = \frac{p(x) }{p(c_1)} \cdot \frac{p(c_1,c_2) \cdot p(x\mid c_1, c_2) }{ p(c_2) \cdot p(x \mid c_2)}
\end{split}
\label{combine_prob_remove}
\end{equation}
Then, by taking the logarithm and computing the gradient with respect to \( x \), we obtain:
\begin{equation}
\begin{split}
\nabla_x \log p(x \mid c_1) = &\nabla_x \log p(x) \\ 
&+ \big[  \nabla_x \log p(x \mid c_1^{'}) - \nabla_x \log p(x) \big] \\
&- \big[ \nabla_x \log p(x \mid c_2) - \nabla_x \log p(x) \big],
\end{split}
\end{equation}
where $c_1^{'}$ denotes $(c_1, c_2)$, and we derive the guidance noise:
\begin{equation}
    \epsilon_{\text{pred}} = \epsilon(\varnothing) + [\epsilon(c_1^{'})-\epsilon(\varnothing)] - [\epsilon(c_2)-\epsilon(\varnothing)].
    \label{eq:pred_remove}
\end{equation}

\subsection{Semantic Manipulation by Noise Control}

\subsubsection{Problem Formulation}

\textbf{GTF} is a training-free semantic manipulation framework that operates by directly modifying noise vectors to control generation in diffusion-based models.
Consider a source prompt \( P_{\text{src}} \) and a target prompt \( P_{\text{tgt}} \), where the latter is derived from the former via the addition, removal, or substitution of specific semantic elements (e.g., converting “a cat” into “a cat wearing sunglasses”). 
To enable controllable manipulation, a guidance vector \( \epsilon_{\text{pred}} \) is synthesized through a geometric composition of the noise estimates associated with the source and target prompts:
\begin{equation}
\begin{split}
    &\epsilon_{\text{pred}} = w_1 \cdot \Psi_1(\epsilon_{\text{src}}) + w_2 \cdot \Psi_2(\epsilon_{\text{tgt}}),\\
    &\epsilon_{\text{src}} = \epsilon_\theta(x_t, P_{\text{src}}) - \epsilon_\theta(x_t, \varnothing ),\\
    &\epsilon_{\text{tgt}} = \epsilon_\theta(x_t, P_{\text{tgt}})- \epsilon_\theta(x_t, \varnothing ),
\end{split}  
\end{equation}
where \( \Psi(\cdot) \) denotes the geometric transformation of noise, \( \epsilon_\theta(x_t, \varnothing) \) is the unconditional noise prediction, and \( w_1 \), \( w_2 \) are scalar weights that modulate the influence of source and target semantics, respectively.  
In the following, we detail how these semantic noise are selectively modulated to enable fine-grained manipulation.

\subsubsection{Generalized and Training-Free Semantic Manipulation}
Text-guided semantic manipulation can be broadly categorized into two types of transformations: \emph{semantic addition} and \emph{semantic removal}.  
As discussed in \cref{sec:explore}, semantic manipulation can theoretically be realized by adjusting the conditional noise in Diffusion models. Building upon this insight, we further devise noise composition strategies tailored to two types of semantic manipulations.

\noindent\textbf{Semantic Addition.}
When the target prompt introduces additional semantics beyond the source (e.g., ``Captain America ...'' becomes ``Captain America ... wearing business suit''), the two prompts share a common semantic core. To isolate the novel attributes introduced by the target, we treat \( \epsilon_{\text{src}} \) as the reference direction and decompose \( \epsilon_{\text{tgt}} \) into parallel and orthogonal components:

\[
\epsilon_{\text{tgt}}^{\parallel} = \frac{\langle \epsilon_{\text{tgt}}, \epsilon_{\text{src}} \rangle}{\| \epsilon_{\text{src}} \|^2} \cdot \epsilon_{\text{src}}, \quad
\epsilon_{\text{tgt}}^{\perp} = \epsilon_{\text{tgt}} - \epsilon_{\text{tgt}}^{\parallel}
\]

Here, the orthogonal component \( \epsilon_{\text{tgt}}^{\perp} \) captures the semantics unique to the target prompt. Then, following the interpretation in \cref{eq:pred_add}, the final guidance noise is formulated as a weighted combination that preserves the shared content while introducing the new semantics:
\[
\hat{\epsilon}_{\text{add}} = w_1 \cdot \epsilon_{\text{src}} + w_2 \cdot \epsilon_{\text{tgt}}^{\perp},
\]

where \( w_1 \) and \( w_2 \) balance the preservation of source semantics and the injection of new attributes.

\noindent\textbf{Semantic Removal.}

When the editing objective involves removing specific semantics from the source prompt (e.g., deleting ``a cup of coffee'' from ``Captain America drinking coffee ...''), the target prompt explicitly specifies the content to be removed (e.g., ``coffee''). In this case, we treat \( \epsilon_{\text{tgt}} \) as the reference direction and decompose the source noise \( \epsilon_{\text{src}} \) accordingly.
\[
\epsilon_{\text{src}}^{\parallel} = \frac{\langle \epsilon_{\text{src}}, \epsilon_{\text{tgt}} \rangle}{\| \epsilon_{\text{tgt}} \|^2} \cdot \epsilon_{\text{tgt}}, \quad
\epsilon_{\text{src}}^{\perp} = \epsilon_{\text{src}} - \epsilon_{\text{src}}^{\parallel}
\]
The projection of \( \epsilon_{\text{src}} \) onto \( \epsilon_{\text{tgt}} \) captures the shared semantics—i.e., the content to be removed. To isolate the remaining information, we discard the parallel component and retain only the orthogonal one. 
Additionally, we subtract the parallel component of source noise vector to derive the final guidance according to \cref{eq:pred_remove}:

\[
\hat{\epsilon}_{\text{remove}} = w_1 \cdot \epsilon_{\text{src}}^{\perp} - w_2 \cdot \epsilon_{\text{src}}^{\parallel},
\]

where the first term preserves the residual content and the second term enforces semantic erasure.
Note that we subtract a scaled version of \( \epsilon_{\text{src}}^{\parallel} \) instead of directly using \( \epsilon_{\text{tgt}} \), since both vectors are aligned in the same semantic direction associated with the target prompt. This choice enables a more interpretable and consistent control over the strength of semantic erasure, as \( w_2 \) smoothly modulates the influence of the shared semantic direction. 

\begin{figure*}
    \centering
    \includegraphics[width=0.95\linewidth]{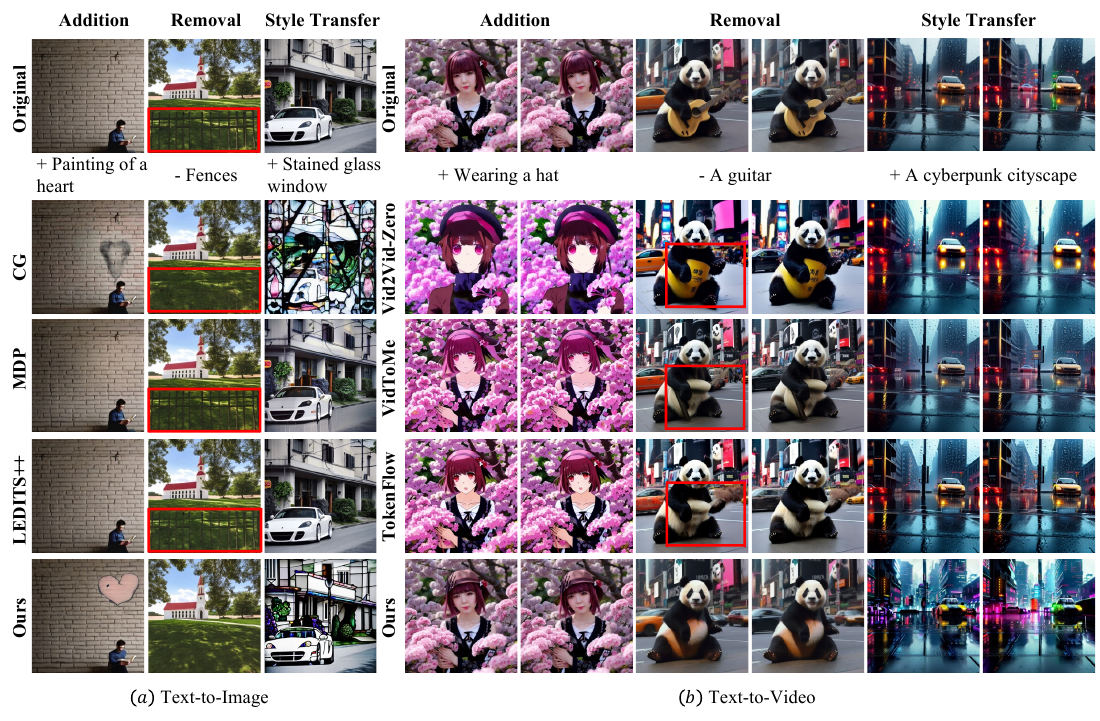}
    \vspace{-12pt}
    \caption{\textbf{Qualitative Comparison with baseline methods.} It demonstrates that our method enables more precise and powerful semantic manipulation while preserving visual consistency. Moreover, it exhibits strong generalization across different modalities and tasks. The source prompts (from left to right) are ``A man reading books in front of a wall", ``A church in the countryside surrounded by fences", ``A sports car driving down the street", ``Sakura Matou surrounded by flowers, anime", ``A panda is playing guitar on times square", and ``A rainy street with cars reflecting on the wet pavement". For each video case, the left and right columns represent the first and last frames, respectively. More results are provided in \cref{app:additional_experiments}.}
    \label{fig:comparison}
\end{figure*}

\section{Experiments}

\subsection{Experiment Setups}

\noindent\textbf{Implementation Details.}
To validate the effectiveness and generalizability of our proposed algorithm, we apply GTF to image, video, and 3D generation tasks across representative diffusion-based frameworks. For image editing, we adopt Stable Diffusion 2.1 (SD)~\cite{Rombach_2022_CVPR} 
with the default PNDM sampler~\cite{liu2022pseudonumericalmethodsdiffusion}. 
For video editing, we intergrate GTF into AnimateDiff~\cite{guo2023animatediff} using the pre-trained motion adapter and Realistic Vision V5.1 model. 
The number of inference steps is set to $ 50 $, with a classifier-free guidance (CFG) scale of $ 7.5 $. Both image and video resolutions are set to $512 \times 512$, with videos consisting of $ 8 $ frames.
For 3D editing tasks, we implement GTF in both LGM~\cite{tang2024lgmlargemultiviewgaussian} and LucidDreamer~\cite{liang2023luciddreamerhighfidelitytextto3dgeneration}, covering general and per-scene pipelines, respectively.  Specifically, we use the default MVDream~\cite{shi2024mvdreammultiviewdiffusion3d} checkpoint with a CFG scale of $ 7.5 $ for LGM; and the same SD backbone as its distillation source and follow the default $ 5000 $ training iterations for LucidDreamer. 
All integrations are training-free and require no additional parameters, and all inference experiments are conducted on a single NVIDIA GeForce RTX 4090 GPU. 
More implementation details can be found in \cref{app:impl_details}.

\noindent\textbf{Dataset.}
We primarily adapt our prompt dataset from PIE-Bench~\cite{ju2023direct}. However, existing prompt datasets often fail to satisfy the diverse input requirements of different editing methods. For instance, MasaCtrl~\cite{cao_2023_masactrl} requires a target prompt that directly describes the desired outcome, while LEDITS++~\cite{brack2024leditslimitlessimageediting} uses a prompt that explicitly specifies what semantic content to add or remove.
To address this, we curate a more versatile dataset in which each sample includes three types of prompts: 
(1) a \textit{source prompt} ($ P_{\text{src}} $) describing the original image or video, 
(2) a \textit{default target prompt} ($ P_{\text{edit}} $) depicting the expected appearance after editing, and 
(3) a \textit{change prompt} explicitly stating the intended semantic modification (addition or removal).

\noindent\textbf{Evaluation Metrics.}
We select three evaluation metrics to comprehensively assess the effectiveness of semantic manipulation, including:
1)  $ \text{CLIP}_{Sim} $,  for semantic addition, we compute CLIP score~\cite{hessel2022clipscorereferencefreeevaluationmetric}, i.e. the cosine similarity between CLIP embeddings of edited image and \textit{default} target prompt to measure how well they align. 2) $ \text{CLIP}_{Inv} $, for semantic removal, we propose to calculate the CLIP score between edited image and \textit{change} prompt to measure how successfully the specified semantics are removed.
3) $ \text{CLIP}_{Dir} $, we also compute CLIP directional similarity~\cite{gal2021stylegannada}, i.e. the cosine similarity between the change from source image to edited image and the change from source prompt to \textit{default} target prompt in CLIP embedding space, to measure how faithful the editing direction is to the semantic change.

\noindent\textbf{User Study.}
We randomly select visual manipulation results based on prompts from our curated dataset for user testing, and compare \methodname\ with other baselines. The participants are asked to consider \textit{alignment of the edited images to the target prompts} (Text Align.), \textit{content consistency of unedited regions of the original images} (Cont. Cons.), \textit{aesthetic appeal that they prefer} (User Pref.), and choose the best edited results for each dimension. Note that we shuffle the options randomly in our questionnaire and do not reveal the methods that the options belong to. More user study details can be found in \cref{app:user_study_setups}

\vspace{-6pt}
\subsection{Comparisons with Text-to-Image Manipulation Methods}
To confirm that our GTF achieves strong semantic manipulation capability, 
we compare it with six SOTA training-free image editing baselines: 1) Prompt-to-Prompt (P2P)~\cite{hertz2022prompt} which utilizes cross-attention layers to control the spatial layout of the image; 2) MasaCtrl~\cite{cao_2023_masactrl} which uses mutual self-attention to query source image; 3) SEGA~\cite{brack2023segainstructingtexttoimagemodels} and 4) MDP~\cite{wang2023mdpgeneralizedframeworktextguided} that manipulate predicted noises; 5) LEDITS++~\cite{brack2024leditslimitlessimageediting} which further combines implicit masking technique; 6) Contrastive Guidance (CG)~\cite{wu2024contrastivepromptsimprovedisentanglement} which achieves editing through disentangling image factors.

\noindent\textbf{Qualitative Comparisons.}
To provide a more intuitive comparison, we visualize qualitative comparison with the more recent and effective baselines in the leftmost three columns in \cref{fig:comparison}. By accurately manipulating noise predictions, our method is capable of adding or removing objects that may occupy a large spatial area, which is often challenging for existing methods. For example, in the second column, the fences are successfully removed only by our method, while other baselines struggle to achieve the desired edit. 
Moreover, our method can effectively introduce the desired style while faithfully preserving the original content. In contrast, other baselines either fail to transfer the target style (e.g., MDP barely shows any visible style change), or overly alter the original appearance (e.g., CG severely distorts the original content).

\noindent\textbf{Quantitative Comparisons.}
For systematic evaluation, we also perform quantitative comparisons against the SOTA image editing baselines in \cref{tab:quantitative_results_image}.
From the table, we can observe the following: 
\romannumeral1) Our method achieves the best results across all metrics, demonstrating its strong capability in semantic manipulation across various tasks; 
\romannumeral2) GTF significantly outperforms all baselines in $ \text{CLIP}_{Dir} $ (36.83), highlighting its superior faithfulness to the intended editing direction; 
\romannumeral3) While CG achieves scores second only to ours, it exhibits noticeable inconsistency with original images, as shown in \cref{fig:comparison}, further underscoring the superiority of our method.

\begin{table}
  \centering
  \caption{\textbf{Quantitative Comparisons with Text-to-Image Manipulation methods.} We highlight the best performance in \textbf{bold} and \underline{underline} the second best.}
  \vspace{-10pt}
  \label{tab:quantitative_results_image}
  \resizebox{\columnwidth}{!}{
  \begin{tabular}{@{}lcccccc@{}}
    \toprule
    Method 
    & \makecell[c]{$\text{CLIP}_{Sim}\uparrow$ \\ (add)} & \makecell[c]{$\text{CLIP}_{Inv} \downarrow$ \\ (remove)} 
    & \makecell[c]{$\text{CLIP}_{Sim}\uparrow$ \\ (style)} 
    & \makecell[c]{$\text{CLIP}_{Dir} \uparrow$ \\ (add)} 
    & \makecell[c]{$\text{CLIP}_{Dir} \uparrow$ \\ (remove)}
    & \makecell[c]{$\text{CLIP}_{Dir} \uparrow$ \\ (style)} \\
    \midrule
    P2P~\cite{hertz2022prompt} & 32.8393 & 21.5167 & \underline{34.8041} & 0.1814 & 0.1408 & 0.1549 \\
    MasaCtrl~\cite{cao_2023_masactrl} & 29.3862 & 21.9416 & 28.5651 & 0.0934 & 0.0803 & 0.0575 \\
    SEGA~\cite{brack2023segainstructingtexttoimagemodels} & 33.4652 & 22.9387 & 31.7848 & 0.1460 & 0.0719 & 0.1253 \\
    MDP~\cite{wang2023mdpgeneralizedframeworktextguided} & 32.8925 & 21.0920 & 34.0180 & 0.1973 & 0.1585 & 0.1516 \\
    CG~\cite{wu2024contrastivepromptsimprovedisentanglement} & \underline{33.8735} & \underline{20.6206} & 34.5988 & \underline{0.2201} & \underline{0.1782} & \underline{0.1696} \\
    LEDITS++~\cite{brack2024leditslimitlessimageediting} & 32.6331 & 21.9653 & 29.7633 & 0.1240 & 0.1313 & 0.0384 \\
    \cellcolor{green!10} GTF (Ours) & \cellcolor{green!10} \textbf{34.8568} & \cellcolor{green!10} \textbf{20.0225} & \cellcolor{green!10} \textbf{36.8254} & \cellcolor{green!10} \textbf{0.2697} & \cellcolor{green!10} \textbf{0.1915} & \cellcolor{green!10} \textbf{0.2349} \\
    \bottomrule
  \end{tabular}
  }
  \vspace{-8pt}
\end{table}

\begin{table}
  \centering
  \caption{\textbf{User Study of GTF and Text-to-Image Manipulation methods.}}
  \vspace{-10pt}
  \label{tab:user_study_image}
  \resizebox{\columnwidth}{!}{
  \begin{tabular}{l|ccccccc}
        \hline
        \diagbox{Metrics}{Method} 
        & P2P & MasaCtrl & SEGA & MDP & CG & LEDITS++ & GTF \\
        \hline
        Text Align. & 8.74\% & 4.83\% & 2.45\% & 15.17\% & 20.14\% & 14.62\% & \cellcolor{pink!20} \textbf{34.06\%} \\
        Cont. Cons. & 7.9\% & 5.52\% & 2.31\% & 15.59\% & 20.49\% & 19.79\% & \cellcolor{pink!20} \textbf{28.39\%} \\
        User Pref. & 8.46\% & 4.20\% & 3.01\% & 16.01\% & 20.91\% & 14.83\% & \cellcolor{pink!20} \textbf{32.59\%} \\
        \hline
      \end{tabular}
  }
  \vspace{-16pt}
\end{table}

\noindent\textbf{User Study.}
We conduct a user study to further validate the effectiveness of our GTF. As can be seen from \cref{tab:user_study_image}, our method receives the highest ratings across all three aspects, confirming its effectiveness in semantic manipulation while maintaining the integrity of the original visual content.

\subsection{Comparisons with Text-to-Video Manipulation Methods}
To further demonstrate the effectiveness of our method in video modality, we also compare it with six SOTA training-free video editing baselines: 1) FateZero~\cite{qi2023fatezero} which fuses attention maps in both inversion and generation processes for temporally consistent video editing; 2) Vid2Vid-Zero~\cite{wang2024zeroshotvideoeditingusing} which uses a spatial-temporal attention module to achieve bi-directional temporal modeling for editing; 3) Rerender-A-Video~\cite{yang2023rerendervideozeroshottextguided} which leverages an adapted diffusion model (DM) to generate key frames and then propagates them to other frames; 4) FLATTEN~\cite{cong2024flattenopticalflowguidedattention} which introduces optical flow into the attention module for visually consistent editing; 5) TokenFlow~\cite{geyer2023tokenflowconsistentdiffusionfeatures} which propagates diffusion features based on inter-frame correspondences; 6) VidToMe~\cite{li2023vidtomevideotokenmerging} which merges self-attention tokens across frames to edit videos. The image editing method that TokenFlow uses is PnP~\cite{tumanyan2022plugandplaydiffusionfeaturestextdriven}.

\begin{table}
  \centering
  \caption{\textbf{Quantitative Comparisons with Text-to-Video Manipulation methods.} We highlight the best performance in \textbf{bold} and \underline{underline} the second best.}
  \vspace{-10pt}
  \label{tab:quantitative_results_video}
  \resizebox{\columnwidth}{!}{
  \begin{tabular}{@{}lcccccc@{}}
    \toprule
    Method 
    & \makecell[c]{$\text{CLIP}_{Sim}\uparrow$ \\ (add)} & \makecell[c]{$\text{CLIP}_{Inv} \downarrow$ \\ (remove)} 
    & \makecell[c]{$\text{CLIP}_{Sim}\uparrow$ \\ (style)} 
    & \makecell[c]{$\text{CLIP}_{Dir} \uparrow$ \\ (add)} 
    & \makecell[c]{$\text{CLIP}_{Dir} \uparrow$ \\ (remove)}
    & \makecell[c]{$\text{CLIP}_{Dir} \uparrow$ \\ (style)} \\
    \midrule
    FateZero~\cite{qi2023fatezero} & 32.5067 & 23.5750 & 34.4693 & 0.0494 & 0.0515 & 0.0850 \\
    Vid2Vid-Zero~\cite{wang2024zeroshotvideoeditingusing} & \underline{33.5822} & 21.6772 & 34.7971 & \underline{0.1152} & 0.1104 & 0.0750 \\
    Rerender-A-Video~\cite{yang2023rerendervideozeroshottextguided} & 32.4598 & 22.1788 & 35.6452 & 0.1062 & 0.0927 & 0.1049 \\
    FLATTEN~\cite{cong2024flattenopticalflowguidedattention} & 32.4373 & 22.9239 & 36.454 & 0.0593 & 0.0590 & \underline{0.1480} \\
    TokenFlow~\cite{geyer2023tokenflowconsistentdiffusionfeatures} & 32.5426 & \underline{21.1212} & 36.0303 & 0.0954 & 0.1053 & 0.1292 \\
    VidToMe~\cite{li2023vidtomevideotokenmerging} & 32.5606 & 21.1732 & \textbf{36.7255} & 0.0819 & \underline{0.1186} & 0.1276 \\
    \cellcolor{green!10} GTF (Ours) & \cellcolor{green!10} \textbf{35.7504} & \cellcolor{green!10} \textbf{20.0050} & \cellcolor{green!10} \underline{36.5175} & \cellcolor{green!10} \textbf{0.2834} & \cellcolor{green!10} \textbf{0.2243} & \cellcolor{green!10} \textbf{0.2143} \\
    \bottomrule
  \end{tabular}
  }
  \vspace{-8pt}
\end{table}

\begin{table}
  \centering
  \caption{\textbf{User Study of GTF and Text-to-Video Manipulation methods.}}
  \vspace{-10pt}
  \label{tab:user_study_video}
  \resizebox{\columnwidth}{!}{
  \begin{tabular}{l|ccccccc}
        \hline
        \diagbox{Metrics}{Method} 
        & FateZero & \makecell[c]{Vid2Vid- \\ Zero} & \makecell[c]{Rerender- \\ A-Video} & FLATTEN & TokenFlow & VidToMe & \makecell[c]{GTF} \\
        \hline
        Text Align. & 3.97\% & 6.03\% & 7.78\% & 5.24\% & 4.13\% & 5.40\% & \cellcolor{pink!20} \textbf{67.46\%} \\
        Cont. Cons. & 5.24\% & 6.35\% & 6.51\% & 3.81\% & 5.24\% & 6.98\% & \cellcolor{pink!20} \textbf{65.87\%} \\
        User Pref. & 4.60\% & 6.51\% & 8.10\% & 3.81\% & 4.44\% & 9.21\% & \cellcolor{pink!20} \textbf{63.33\%} \\
        \hline
      \end{tabular}
  }
  \vspace{-10pt}
\end{table}
\noindent\textbf{Qualitative Comparisons.}
To demonstrate the strong editing fidelity of our method, we present a visual comparison with prominent baselines in the rightmost six columns of \cref{fig:comparison}. From the results, we observe that our method (bottom row) not only successfully performs the intended edits but also better preserves the visual characteristics of the source videos. In contrast, other baselines struggle to achieve both objectives simultaneously. 
For object addition, Vid2Vid-Zero manages to insert the ``hat'' but fails to maintain the original content, while VidToMe and TokenFlow do not successfully add the object. For object removal, all three baselines struggle to eliminate the ``guitar'' completely, leading to noticeable artifacts. In the case of style transfer, the baselines fail to convincingly apply the ``Cyberpunk'' style, resulting in suboptimal visual outcomes.

\begin{figure*}[]
    \centering
    \vspace{-10pt}
    \includegraphics[width=\linewidth]{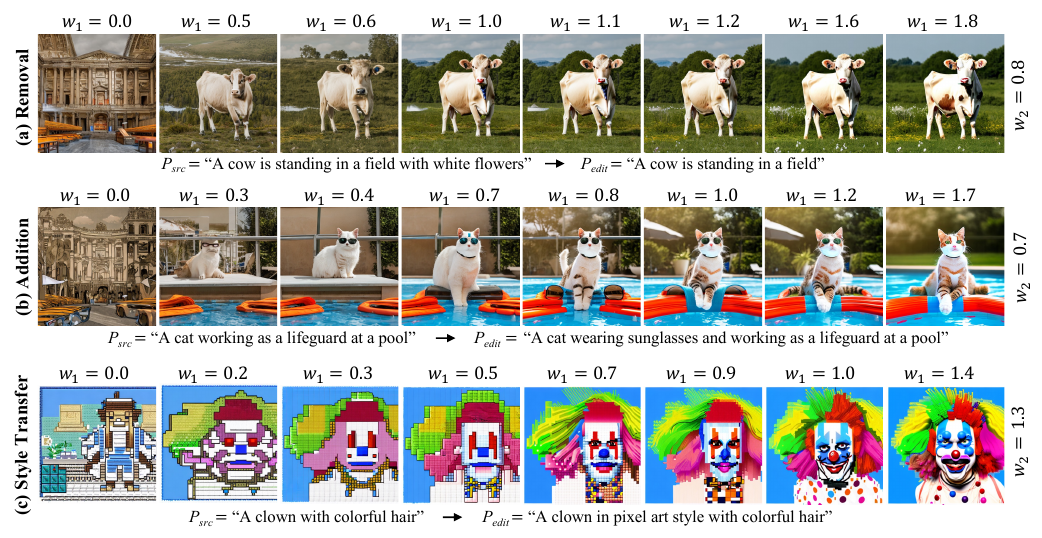}
    \vspace{-20pt}
    \caption{\textbf{Ablation Study on the Impact of $w_1$.} It demonstrates that when $w_1$ is too large, the result is overly influenced by the source prompt, hindering correct semantic manipulation. Conversely, if $w_1$ is too small, the source semantic support becomes insufficient, leading to a deviation from the original semantic structure in the generated content.}
    \label{fig:ablation_w1}
\end{figure*}

\begin{figure*}
    \centering
    \vspace{-10pt}
    \includegraphics[width=\linewidth]{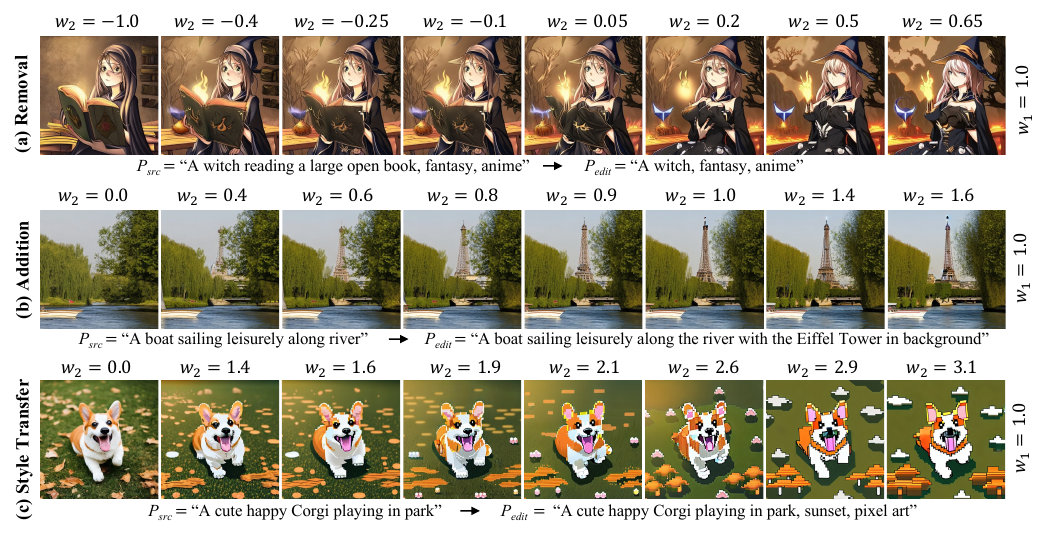}
    \vspace{-18pt}
    \caption{\textbf{Ablation Study on the Impact of $ w_2 $.} We fix $w_1 = 1.0$ and show the \emph{smooth changes} in the generated results while gradually increasing $w_2$. Note that we do not require any mask to restrict the editing area.}
    \label{fig:ablation_w2}
    \vspace{-12pt}
\end{figure*}

\noindent\textbf{Quantitative Comparisons.}
Similarly, we conduct quantitative comparisons between baselines and our GTF in \cref{tab:quantitative_results_video}.
From the table, we observe that: 
\romannumeral1) Our method achieves the best results in five out of six metrics, demonstrating its strong generalization ability and semantic manipulation performance across tasks; 
\romannumeral2) The performance trend is consistent with that in image editing, which indicates the inherent manipulation fidelity of our method regardless of the specific modality; 
\romannumeral3) Although our $ \text{CLIP}_{Sim} $ score in the style transfer task does not reach the highest value, it ranks second with only a marginal gap of 0.2. Moreover, visual comparisons in the \cref{fig:comparison} show that our method significantly outperforms VidToMe in preserving consistency before and after manipulation.

\noindent\textbf{User Study.}
As illustrated in \cref{tab:user_study_video}, our method demonstrates overwhelming superiority over all baselines across the three evaluation dimensions. For example, in \textit{Text Align}, our method achieves a remarkable score of 67.46\%, while the second-best baseline only reaches 7.78\%, highlighting the substantial advantage of our approach in aligning edits with user intent. This also strongly supports the generalizability of our method across different modalities.

\begin{figure}[h]
    \centering
    \includegraphics[width=1\columnwidth]{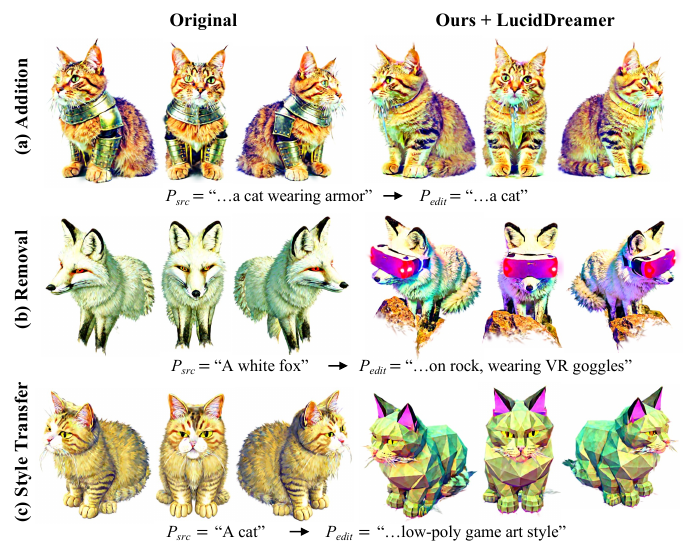}
    \vspace{-10pt}
    \caption{\textbf{Qualitative Results of LucidDreamer Combined with Our GTF.} Note that the fine-grained details of the cat's face are largely preserved, despite the fact that no explicit mask is applied.}
    \vspace{-16pt}
    \label{fig:qualitative_luciddreamer}
\end{figure}

\begin{figure}
    \centering
    \includegraphics[width=\columnwidth]{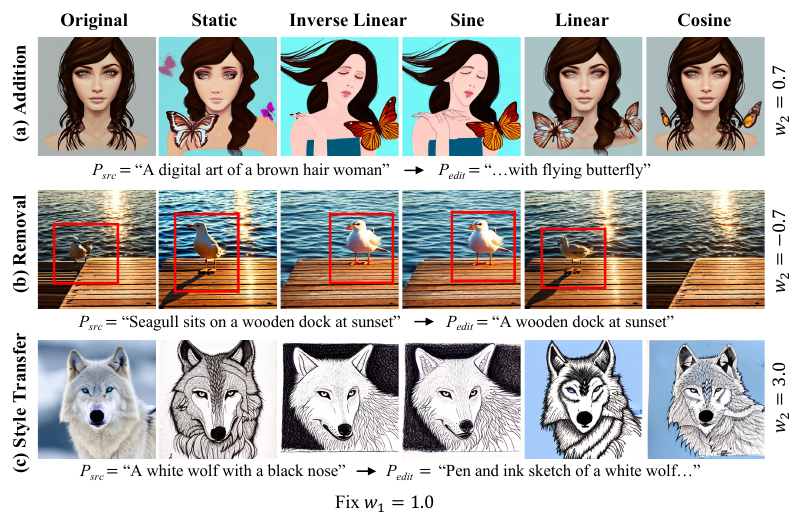}
    \vspace{-18pt}
    \caption{\textbf{Ablation Study on the effect of weight schedulers.} }
    \label{fig:ablation_scheduler}
    \vspace{-12pt}
\end{figure}

\subsection{Extension to Text-to-3D Manipulation}
We also reveal the potential of \methodname\ in 3D editing tasks. For both paradigms of 3D generation — general and per-scene — we successfully integrate our training-free GTF method. To intuitively demonstrate its effectiveness, we visualize the editing results for both settings. In \cref{fig:teaser}(c), we show results for the general paradigm, where our method better preserves the consistency of the original 3D content. In \cref{fig:qualitative_luciddreamer}, we present results for the per-scene paradigm, where our method not only ensures high-fidelity visual outputs but also achieves accurate semantic manipulation. These results further demonstrate the strong generalizability of GTF across both tasks and modalities. For more implementation details please refer to \cref{app:impl_details}.

\subsection{Ablation Studies}

\noindent\textbf{Effect of Weight Schedulers.}
To further investigate the influence of diverse weight schedulers on semantic manipulation, we conducted an in-depth validation process, utilizing various scheduler configurations as depicted in \cref{fig:ablation_scheduler}.
From the visualization results, we find that: 1) For \textit{static} scheduler (i.e. $ w_2 $ remains constant throughout generation process), although edited results could more adhere to newly injected semantics (e.g. the drawing of the wolf at the bottom row becomes entirely black and white to adhere to ``pen and ink sketch"), characteristics of the source image could be significantly altered; 2) For monotonically decreasing schedulers (i.e. \textit{inverse linear} and \textit{sine} schedulers), the edited results could severely deviate from the source image; 3) We find the monotonically increasing schedulers (i.e. \textit{linear} and \textit{cosine} schedulers) to be more optimal among the five weight schedulers that we experiment.
However, cosine scheduler is generally more capable of preserving original content and thus can achieve desired edit without influencing irrelevant regions of the source image (see how the ``seagull" is successfully removed by cosine scheduler while linear scheduler fails to, and how linear scheduler alters the girl's hair when adding butterflies next to her). We attribute the phenomena mentioned above to the empirically discovered nature of diffusion-based generation process, i.e. the early stages of generation determine general layout while the later stages refine the image details. Consequently, monotonically decreasing schedulers will produce results that severely differ from the source images due to its larger $ w_2 $ at the early stages; Meanwhile, since cosine scheduler injects the weakest $ w_2 $ in early stages and the strongest in later stages, it can achieve desired edits with minimal influence on unedited regions.

\noindent\textbf{Effect of $ w_1 $.}
Next, we explore the effect of $ w_1 $. Because $w_1$ controls the strength of semantics from source prompts, altering $w_1$ has a similar effect as changing CFG scale. As shown in \cref{fig:ablation_w1}, when $w_1$ is too low, guidance from source prompts will not be enough to generate reasonable object shapes. Meanwhile, as $w_1$ increases and semantics from source prompts become stronger, we need to strengthen the semantics from target prompts by increasing $w_2$ accordingly to achieve the same editing effects. As shown in the rightmost column of \cref{fig:ablation_w1}, because the strengths of the source semantics are much higher than those of the target, desired edits are not achieved.

\noindent\textbf{Effect of $ w_2 $.}
Finally, we ablate the effect of $ w_2 $. As shown in \cref{fig:ablation_w2} $(a)$, as $w_2$ increases in semantic removal, the target object (``a large open book") gets thinner until it vanishes when $w_2$ is around $0.05$. The meaning of $w_2$ is more straightforward in semantic addition. The higher $w_2$ is, the more semantics will be injected into the original image, resulting in clearer and larger shapes (e.g. the tower in $(b)$), and stronger artistic styles (e.g. the ``resolution" in $(c)$ getting lower to conform to ``pixel art"), etc.

\section{Conclusion}
In this work, we first investigate the geometric properties of noise in diffusion models and reveal their strong correlation with semantic transformations. Building upon this insight, we propose \emph{\methodname}, a novel training-free framework for text-guided semantic manipulation. Our method enables flexible and efficient editing — including addition, removal, and style transfer — without requiring model tuning. Extensive experiments demonstrate that \emph{\methodname} delivers high-fidelity results with strong generalizability across modalities and tasks, advancing the state-of-the-art in semantic manipulation.

\bibliographystyle{ACM-Reference-Format}
\bibliography{main}

\appendix
\setcounter{page}{1}

\section{Related Works}

\subsection{Text-to-Image}

Text-to-image editing has emerged as a highly active research area. Early work centered on image synthesis and editing using Generative Adversarial Networks (GANs) \cite{dhamo2020semanticimagemanipulationusing, li2020lightweightgenerativeadversarialnetworks, lang2021explainingstyletraininggan, tao2022denetdynamictextguidedimage}, encompassing both conditional \cite{Frolov_2021} and unconditional GANs \cite{karras2019stylebasedgeneratorarchitecturegenerative, xia2022ganinversionsurvey}. However, challenges such as mode collapse and limited cross-modal consistency have prompted extensive exploration into diffusion models \cite{song2022denoisingdiffusionimplicitmodels, ho2020denoisingdiffusionprobabilisticmodels}. These models \cite{Avrahami_2022, gal2022imageworthwordpersonalizing, nichol2022glidephotorealisticimagegeneration, zhang2023addingconditionalcontroltexttoimage} provide a unified framework for managing multimodal data and have shown robust performance across diverse datasets. For instance, SDEdit~\cite{meng2022sdeditguidedimagesynthesis} applies noise to an image and then employs a stochastic differential equation (SDE) prior to iteratively denoise and synthesize realistic images. DiffusionCLIP~\cite{kim2022diffusioncliptextguideddiffusionmodels} combines a diffusion model with CLIP guidance, enabling controlled manipulation of image attributes associated with text, thus facilitating the generation of new images within stylistically related domains. DreamBooth~\cite{ruiz2023dreamboothfinetuningtexttoimage} and Textual Inversion~\cite{gal2022imageworthwordpersonalizing} generate novel images by combining multiple input images with guiding textual descriptions. Imagic~\cite{kawar2023imagictextbasedrealimage}, in contrast, refines specific regions within an image by fine-tuning a pre-trained diffusion model, thereby allowing targeted, text-driven editing tasks on a single image.

\subsection{Text-to-Video}

Compared to text-guided image editing, text-driven video editing presents greater challenges due to the need to maintain temporal consistency both within and outside the editing scope. Early approaches \cite{esser2023structurecontentguidedvideosynthesis, molad2023dreamixvideodiffusionmodels} trained on large-scale text-video datasets to develop models capable of text-guided editing while preserving overall coherence; however, the high computational cost of these methods limits their accessibility for most researchers. To address this issue, some approaches \cite{chen2024controlavideocontrollabletexttovideodiffusion, yan2023magicpropdiffusionbasedvideoediting} utilize pre-trained text-to-video (T2V) models for video editing. Pix2Video~\cite{ceylan2023pix2videovideoeditingusing}, for instance, applies a diffusion model to edit key frames based on textual input, subsequently propagating these edits to other frames via self-attention features. Vid2vid-zero~\cite{wang2024zeroshotvideoeditingusing} achieves temporal and spatial consistency by employing cross-frame temporal modeling and spatial regularization of the original video. However, these approaches often apply edits across entire frames \cite{ceylan2023pix2videovideoeditingusing, wu2023tuneavideooneshottuningimage}, which can lead to unintended alterations in non-edited areas. To address this limitation, VidEdit~\cite{couairon2024videditzeroshotspatiallyaware} integrates a panoramic segmenter and edge detector, enabling object-level control over edited content. Edit-A-Video~\cite{shin2023editavideosinglevideoediting} further refines editing precision by using an attention map to target specific regions. Additionally, MeDM~\cite{chu2023medmmediatingimagediffusion} and FLATTEN~\cite{cong2024flattenopticalflowguidedattention} incorporate optical flow into the diffusion model, effectively preserving temporal consistency and minimizing redundant content across frames during video editing.

\subsection{Text-to-3D}

Significant progress has been made in the field of text-to-3D editing in recent years. Early work primarily focused on image-centered generation and editing techniques \cite{Oppenlaender_2022, li2024instructpix2nerfinstructed3dportrait}, introducing image-editing approaches to 3D editing \cite{haque2023instructnerf2nerfediting3dscenes, kamata2023instruct3dto3dtextinstruction} by reconstructing 3D models through edits on 3D renderings. Building on this, various enhancement methods, such as multi-view consistent editing \cite{dong2024vicanerfviewconsistencyaware3dediting, song2023blendingnerftextdrivenlocalizedediting, mirzaei2023watchstepslocalimage} and general-purpose editing \cite{khalid2024latenteditortextdrivenlocal, chen2023shapeditorinstructionguidedlatent3d, fang2023editing3dscenestext}, have been developed to improve the quality and efficiency of 3D editing.
Moreover, diffusion model-based approaches in the text-to-3D domain \cite{poole2022dreamfusiontextto3dusing2d} have gained considerable attention. By using pre-trained text-to-image diffusion models, these methods generate target data through an iteratively noising and denoising process, showcasing the potential of diffusion models in 3D editing \cite{chen2023shapeditorinstructionguidedlatent3d, kamata2023instruct3dto3dtextinstruction}. To address the multi-view consistency challenges that arise from editing 3D renderings, several studies \cite{hyung2023local3dediting3d, memery2023generatingparametricbrdfsnatural, ma2023xmeshfastaccuratetextdriven} have leveraged text-vision models \cite{poole2022dreamfusiontextto3dusing2d} by connecting the edited 3D models to powerful pre-trained text-vision models, further exploring text-driven 3D editing. Meanwhile, thanks to advances in Score Distillation Sampling (SDS), recent works on 3D editing of highly detailed objects and complex scenes \cite{cheng2024progressive3dprogressivelylocalediting, mikaeili2023skedsketchguidedtextbased3d, decatur20233dpaintbrushlocalstylization} have also achieved impressive results.

\section{Weight Schedulers}

\subsection{Mathematical Definitions}

We employ five different types of weight schedulers in $w_2$ in our experiments: one static, two that increase monotonically, and two that decrease monotonically. They are mathematically defined as follows:
\begin{align}
    \text{static: } w(t) &= w_0 \\
    \text{linear: } w(t) &= w_0 \cdot 2 \cdot (1 - \frac{t}{T}) \\
    \text{cosine: } w(t) &= w_0 \cdot (\cos{\frac{\pi t}{T}} + 1) \\
    \text{inverse linear: } w(t) &= w_0 \cdot 2 \cdot \frac{t}{T} \\
    \text{sine: } w(t) &= w_0 \cdot [\sin{(\frac{\pi t}{T} - \frac{\pi}{2})} + 1]
\end{align}

Where $T$ is the maximum timestep in diffusion sampling (typically $1000$). Following the practice of Xi Wang et al.~\cite{wang2024analysisclassifierfreeguidanceweight}, we integrate the weight functions over timestep $t$ and normalize them to equal to $w_0 \cdot T$, thus the coefficient $2$ in \textbf{linear} and \textbf{inverse linear} schedulers. We find that monotonically increasing schedulers are consistently better in preserving characteristics of original scene, though there are exceptions where static scheduler may be preferred (as discussed in \cref{app:additional_experiments}). We also find that monotonically decreasing schedulers generally perform worse than other schedulers, which is consistent with our conjecture that stronger injection of target semantics in early stages will lead to more significant alterations of original scene.

\section{Experiment Setups}

\subsection{Implementation Details}
\label{app:impl_details}

The qualitative and quantitative results of the baseline methods are obtained using their official implementations with default hyperparameters, except that we employ the same SD checkpoint to ensure a fair comparison.

\noindent \textbf{Implementation Details of Image Editing Baselines.} We use \textit{change} prompt as target prompt for LEDITS++ following their official examples, and use \textit{default} target prompt for others. We input source prompt for all inversions. Note that since SEGA does not include inversion in their implementation, we first generate its exclusive source image using the same random seed that generates original source image, and we use this exclusive source image to compute $ \text{CLIP}_{Dir} $ for SEGA to ensure fairness. 

\noindent \textbf{Implementation Details of 3D Editing.} There are primarily two types of text-to-3D pipelines, namely the general ones which rely on large-scale pretraining but can output results in a forward pass, and the per-scene ones which leverage off-the-shelf 2D diffusion models but require iterative and time-consuming per-scene generation.
1) For generic text-to-3D manipulation, we apply our method to LGM~\cite{tang2024lgmlargemultiviewgaussian} which utilizes multi-view Gaussian features as 3D representation. More specifically, we first generate the source 3D asset by only conditioning on the source prompt, then we apply \methodname\ to the multi-view DM (MVDream~\cite{shi2024mvdreammultiviewdiffusion3d}) to produce edited multi-view images for LGM to generate edited 3D Gaussian~\cite{kerbl20233dgaussiansplattingrealtime} representation. Note that the diffusion sampling space of multi-view DM could be fundamentally different from 2D DM, which is also the case with AnimateDiff, whose sampling space is specifically trained for text-to-video domain. Therefore, the fact that both models can be integrated with \methodname\ and produce convincing results demonstrate the universal applicability of our method. 2) For per-scene text-to-3D manipulation, we apply our method to LucidDreamer~\cite{liang2023luciddreamerhighfidelitytextto3dgeneration} which bases its generation process on distilling 2D DM. Adapting our method to LucidDreamer is similar to modifying SD for image editing, except that we also combine our method with Perp-Neg~\cite{armandpour2023reimaginenegativepromptalgorithm} since it's used as default by LucidDreamer and it helps to maintain 3D consistency.

\subsection{Supplementary Experiment Results}
\label{app:additional_experiments}

\begin{figure}
    \centering
    \includegraphics[width=\columnwidth]{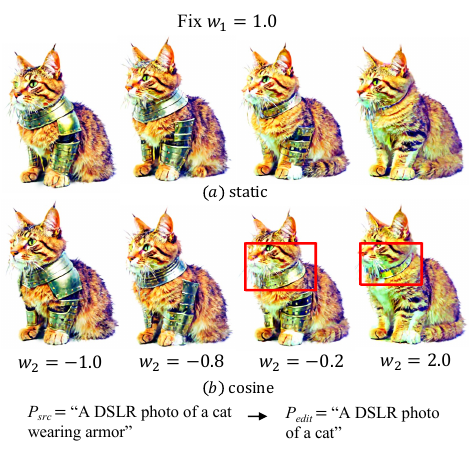}
    \caption{\textbf{Ablation of $w_2$ in text-to-3D removal task.} Although cosine scheduler is better in preserving original characteristics, it could fail to achieve desired results when model's sampling space is too steady.}
    \label{fig:ablation_scheduler_cat}
\end{figure}

\begin{figure}
    \centering
    \includegraphics[width=\columnwidth]{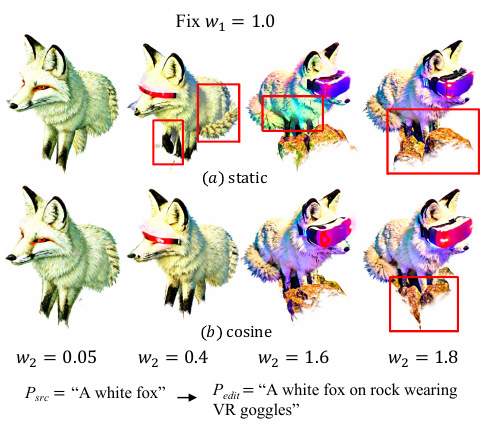}
    \caption{\textbf{Ablation of $w_2$ in text-to-3D addition task.} Compared with cosine scheduler, static scheduler can inject more target semantics into the original scene (e.g. bigger rock in $(a)$ when $w_2 = 1.8$), but also leads to more evident changes in original characteristics (e.g. the pose of the fox in $(a)$ when $w_2 = 0.4$).}
    \label{fig:ablation_scheduler_fox}
\end{figure}

\noindent \textbf{Ablation Study on Weight Schedulers for 3D Editing.}
We show more comparisons between different weight schedulers in \cref{fig:ablation_scheduler_cat} and \cref{fig:ablation_scheduler_fox} for 3D editing tasks. Although cosine scheduler is a good default choice in most scenarios, it could be suboptimal if the sampling space of the model is too stable. We also find this to be more frequent in removal than addition tasks. This phenomenon is illustrated in \cref{fig:ablation_scheduler_cat}. When $w_2 = 2.0$, cosine scheduler fails to completely remove the "armor" around the cat's neck.

\begin{figure*}
    \centering
    \vspace{-28pt}
    \includegraphics[width=\linewidth]{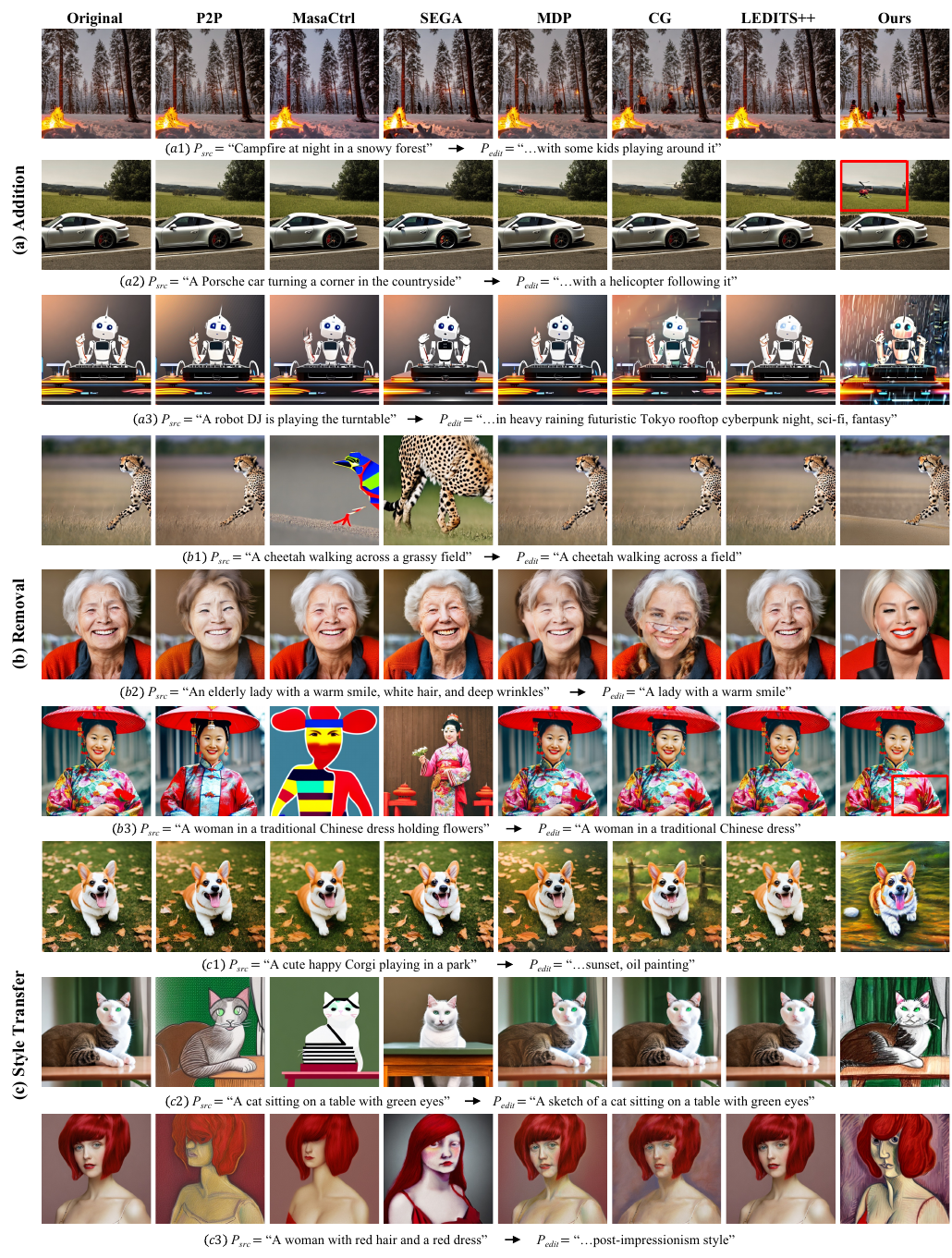}
    \caption{More visual results of our \methodname\ on text-guided image editing.}
    \label{fig:appendix_image}
\end{figure*}

\begin{figure*}
    \centering
    \includegraphics[width=\linewidth]{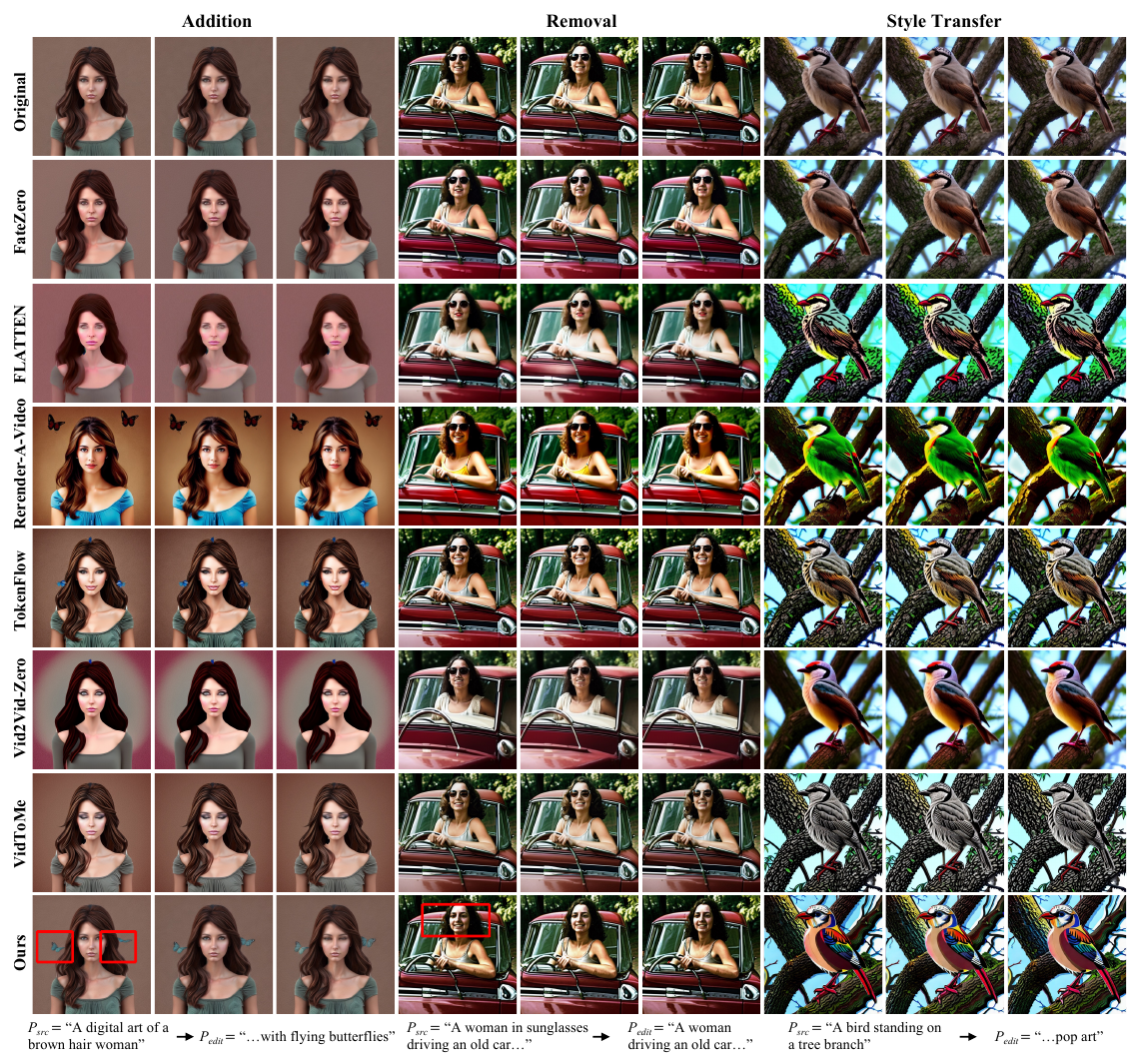}
    \caption{More visual results of our \methodname\ on text-guided video editing.}
    \label{fig:appendix_video}
\end{figure*}

\noindent \textbf{Supplementary Qualitative Results.} We show more visual comparisons between our \methodname\ and other baselines in \cref{fig:appendix_image} and \cref{fig:appendix_video}. We note that existing methods often struggle to simultaneously achieve desired edits and faithfully preserve original characteristics. This is more evidently demonstrated by the semantic addition results of Rerender-A-Video in \cref{fig:appendix_video}. Although it successfully adds ``butterflies'' into the video, the woman's characteristics (e.g. the color of her dress) are significantly changed. The fact that our \methodname\ facilitates both editing faithfulness and content consistency of unedited regions at the same time demonstrates its superiority over existing methods.

\subsection{User Study Setups}
\label{app:user_study_setups}

\noindent \textbf{User Study Details.} For user study on image editing, we collect $ 1430 $ valid votes in total for each measuring dimension; for video editing, the valid votes add up to $ 630 $ for each dimension. All the cases presented in our questionnaires are chosen at random from our curated dataset.

\begin{figure*}
    \centering
    \vspace{-18pt}
    \includegraphics[width=0.7 \linewidth]{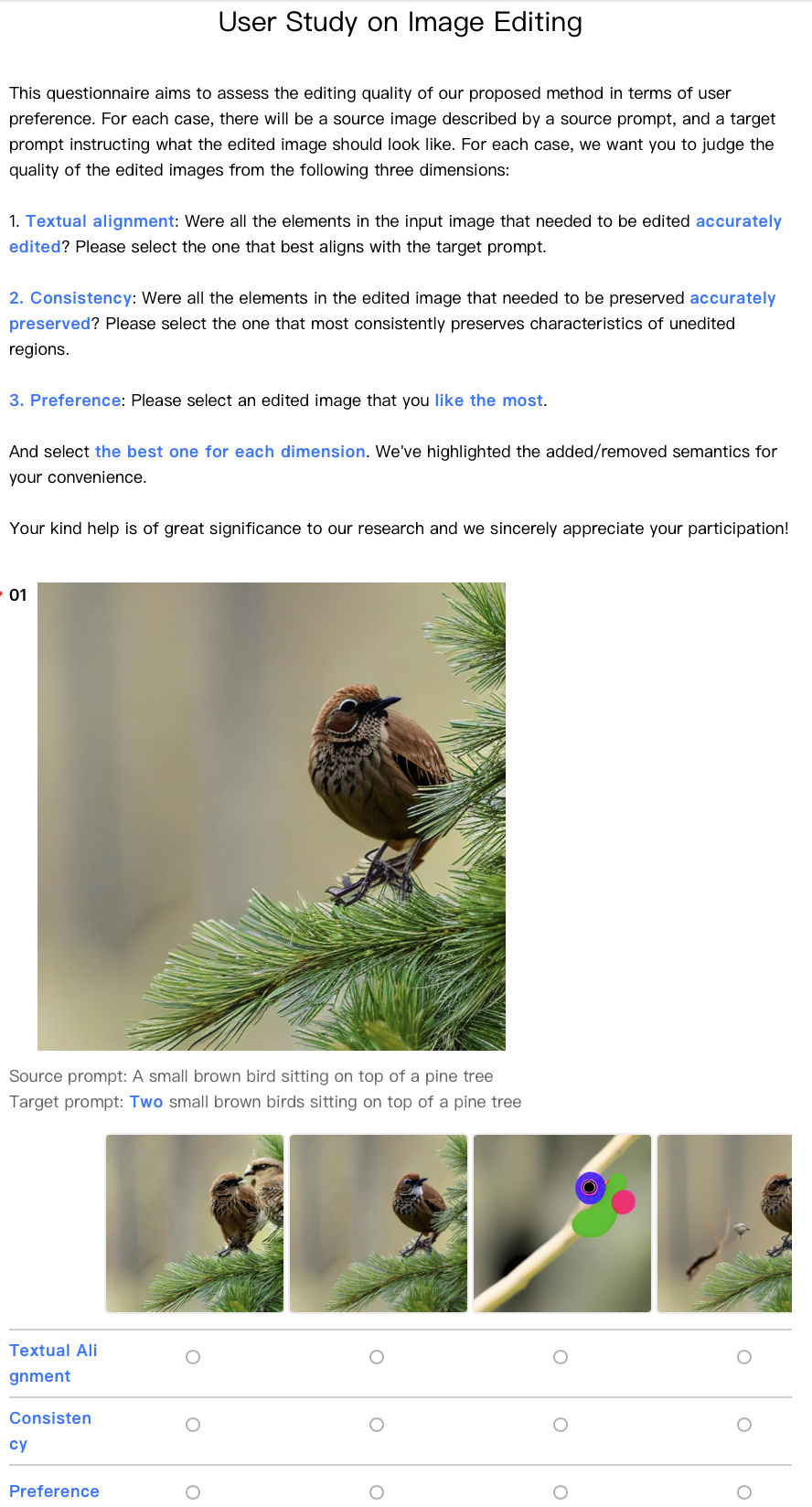}
    \caption{An example of our user study interface for image editing.}
    \label{fig:questionnaire}
\end{figure*}

\noindent \textbf{Questionnaire Interface.} We provide a screenshot of our user study questionnaire on image editing in \cref{fig:questionnaire}. Given a source image, a source prompt, and a target prompt, participants are asked to select their preferred images from six edited results across three dimensions (i.e. textual alignment, consistency, and user preference).
The interface of our questionnaire on video editing is largely the same, except that each edited result is changed from an image to a video represented by the first, the middle, and the last frames.

\end{document}